\documentclass[10pt,twocolumn,letterpaper]{article}

\usepackage{iccv}              

\usepackage{algorithm} 
\usepackage{algorithmic} 

\usepackage{caption} 
\usepackage{amsmath}
\usepackage{amssymb}
\usepackage{mathtools}
\usepackage{amsthm}
\usepackage{multirow}
\usepackage{pifont}
\newtheorem{theorem}{Theorem} 
\newtheorem{definition}{Definition}
\definecolor{iccvblue}{rgb}{0.21,0.49,0.74}
\usepackage[pagebackref,breaklinks,colorlinks,allcolors=iccvblue]{hyperref}

\def\paperID{2242} 
\def\confName{ICCV}
\def\confYear{2025}

\title{Straighten Viscous Rectified Flow via Noise Optimization}

\author{Jimin Dai\textsuperscript{1}, Jiexi Yan\textsuperscript{2}, Jian Yang\textsuperscript{1}, Lei Luo\textsuperscript{1}\\
\textsuperscript{1}PCA Lab, Nanjing University of Science and Technology, \textsuperscript{2}Xidian University\\
{\tt\small \{jimindai, csjyang\}@njust.edu.cn, \{jxyan1995, luoleipitt\}@gmail.com}}

\begin{document}
\maketitle
\begin{abstract}
The Reflow operation aims to straighten the inference trajectories of the rectified flow during training by constructing deterministic couplings between noises and images, thereby improving the quality of generated images in single-step or few-step generation. 
However, we identify critical limitations in Reflow, particularly its inability to rapidly generate high-quality images due to a distribution gap between images in its constructed deterministic couplings and real images.
To address these shortcomings, we propose a novel alternative called Straighten Viscous Rectified Flow via Noise Optimization (VRFNO), which is a joint training framework integrating an encoder and a neural velocity field. 
VRFNO introduces two key innovations: (1) a historical velocity term that enhances trajectory distinction, enabling the model to more accurately predict the velocity of the current trajectory, and (2) the noise optimization through reparameterization to form optimized couplings with real images which are then utilized for training, effectively mitigating errors caused by Reflow's limitations.
Comprehensive experiments on synthetic data and real datasets with varying resolutions show that VRFNO significantly mitigates the limitations of Reflow, achieving state-of-the-art performance in both one-step and few-step generation tasks.
\end{abstract}    
\section{Introduction}
\label{submission}
\setlength{\intextsep}{4pt plus 2pt minus 2pt}
\setlength{\abovecaptionskip}{2pt plus 1pt minus 2pt}
Diffusion models (DMs) \cite{ho2020denoising,song2020denoising,saharia2022palette} have attracted significant attention due to their remarkable generative capabilities and stable optimization process. DMs have outperformed traditional models such as generative adversarial networks (GANs) \cite{adler2018banach,chu2024attack} and Variational autoencoders (VAEs) \cite{yoo2024topic} in various generation tasks. DMs work by designing a forward noise-adding scheme, which progressively transforms the target distribution into the initial distribution (usually a Gaussian distribution), thereby constructing a Probability Flow (PF) from complex to simple distributions. DMs learn the inverse process of PF, and then they can sample Gaussian noises and generate samples that approximate the target distribution through multiple inference steps.
Currently, various forward noise-adding schemes have been developed for DMs. Among them, the most straightforward approach is the Rectified Flow (RF) \cite{liu2022flow,liu2023instaflow,roy20242}. The core idea of RF is to add noise to the data in a linear manner, constructing a deterministic PF from the Gaussian distribution to the target distribution. The model only needs to learn the velocity field of the straight-line interpolation trajectories between the noises and the samples. Once trained, the model can efficiently generate samples that approximate the target distribution.

Generating samples with DMs typically requires multiple inference steps that consume substantial computational resources and time. RF constructs straight-line interpolation trajectories between noises and samples as the reference, and it should theoretically learn a straight flow from the Gaussian distribution to the target distribution. The straight flow implies that RF learns a constant velocity field, which should enable few-step or even single-step sampling in an ideal scenario.
However, during actual training, the RF struggles to learn a constant velocity field, and the resulting PF trajectories remain curved. Therefore, RF still relies on multiple inference steps to generate the desired images. RF attributes the curvature to the crossing of reference trajectories during training and employs a Reflow operation to mitigate this issue.

In this paper, we observe that during the actual training process, the probability of crossing between the straight-line interpolation trajectories of randomly matched noises and images at a given time step $t$ is $P\sim O(e^{-c(n\times n)})$, which is very small in high-dimensional data space (Theorem \ref{Theorem_1}). Therefore, we are more inclined to believe that the curvature arises because the neural network model is not an exact solver, intermediate states along different trajectories may become “approximately crossing” due to their high similarity in statistical property, thereby interfering with the model predictions. Furthermore, we conducted an in-depth analysis of the factors that contribute to the success of Reflow, as well as its limitations, detailed in Section \ref{3.2}. 

To retain the advantages of Reflow while mitigating its drawbacks, we propose a new method called Straighten Viscous Rectified Flow via Noise Optimization (VRFNO). VRFNO introduces a historical velocity term (HVT) as input to the neural velocity field, allowing the model to distinguish the direction of the PF in the presence of approximate crossing. By incorporating historical velocity information, the model can more accurately predict the velocity.
Moreover, VRFNO directly uses the original dataset to avoid the distribution gap and achieves more appropriate matches between the noises and the images  (we refer to it as \textit{optimized coupling}) by noise optimization. Specifically, we propose a joint training framework for an encoder and a neural velocity field. The encoder first encodes images from randomly matched noise-image pairs and then outputs the corresponding mean and variance matrices. The mean and variance matrices are combined with the noise using the reparameterization technique to generate a new noise, which is then fed into the neural velocity field for training.
This joint training approach enables the encoder to adjust the mean and variance of the noise based on the image, thereby optimizing the matches between the noises and the images, which in turn achieves \textit{optimized couplings}. By using these \textit{optimized couplings} to train the velocity field, the model can more efficiently straighten the inference trajectory during training, enabling one-step or few-step generation.

In summary, our contributions are as follows:
\begin{itemize}
  \item[•]We propose a new method VRFNO to straighten the inference trajectories for higher quality few-step and single-step generation.
  \item[•]We introduce an HVT as the auxiliary information to enhance the model's accuracy in velocity prediction when the flow approximates crossing.
  \item[•]We construct \textit{optimized couplings} by optimizing the noise to train the neural velocity field. This avoids relying on \textit{deterministic couplings} used in Reflow, which will constrain the model's generation quality.
\end{itemize}
\section{Preliminary}
\begin{figure}[t!]
\centering
\centerline{\includegraphics[width=\columnwidth]{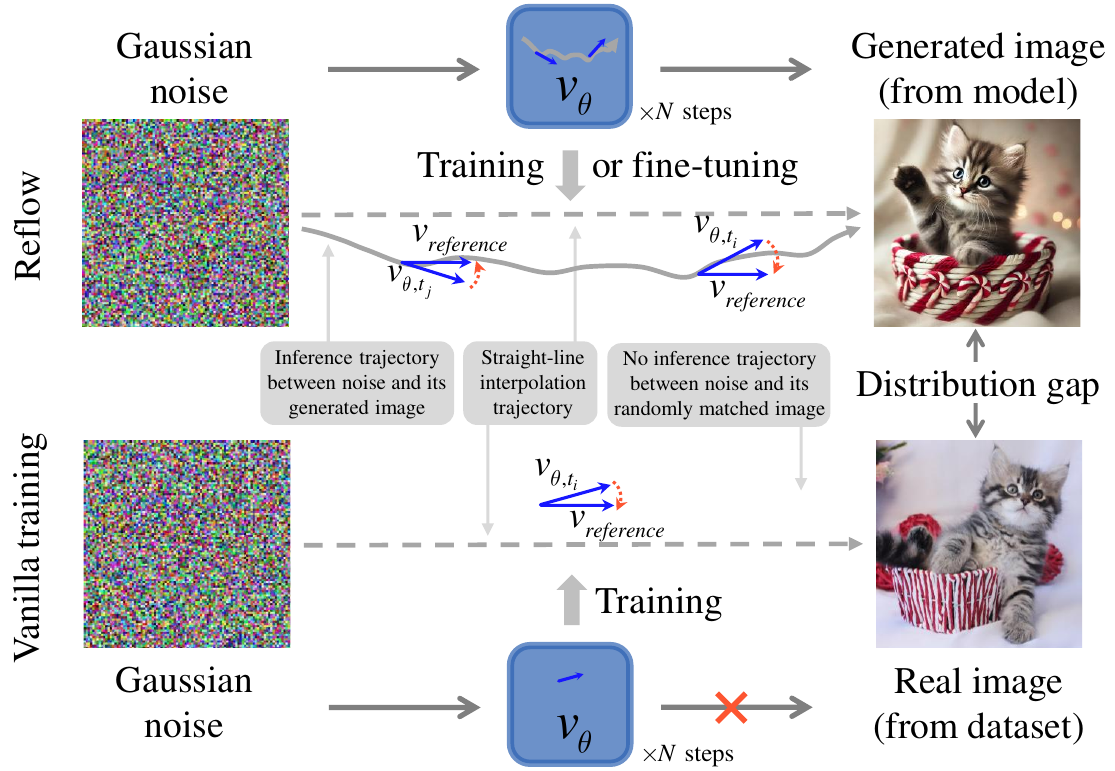}}
\caption{\textbf{Comparison between Reflow and Vanilla training mode.} The top illustrates the Reflow training mode, where noises generate images via a pre-trained model, forming \textit{deterministic couplings} for training. These couplings are reused during training by randomly sampling different intermediate states along the trajectories. The bottom describes the vanilla training mode, where noises and images are randomly sampled to form \textit{arbitrary couplings} for training, without reusing data during training.}
\label{dongji}
\end{figure}

\subsection{Rectified Flow}
RF is an ordinary differential equations (ODE)-based DM that aims to learn the mapping between two distributions, $\pi_0$ and $\pi_1$. In image generation tasks, $\pi_0$ typically represents a standard Gaussian distribution, while $\pi_1$ corresponds to the target image distribution. For empirical observations $X_0 \sim \pi_0$ and  $X_1 \sim \pi_1$ on time $t\in [0,1]$, RF is defined as:
\begin{equation}
d{X_t} = v({X_t},t)dt,
\label{eq:1}
\end{equation}
where intermediate state ${X_t} = t{X_1} + (1 - t){X_0}$ represents a time-differentiable interpolation between $X_0$ and $X_1$, $v:\mathbb{R}^{n\times n} \times [0,1] \to \mathbb{R}^{n \times n}$ denotes the velocity field defined on the data-time domain. Since RF uses simple straight-line interpolation to connect $X_0$ and $X_1$, its trajectory has a constant velocity field ${v_{ref}} = {X_1} - {X_0}$. Thus, the training process optimizes the model by solving a least-squares regression problem, i.e., fitting $v_{ref}$ to neural velocity field $v_\theta$:
\begin{equation}
\min \mathbb{E} {_{{X_0},{X_1}\sim\gamma ,t \in p(t)}}\left[ {{{\left\| {{v_{ref}} - {v_\theta }({X_t},t)} \right\|}^2}} \right],
\label{eq:2}
\end{equation}
where $\gamma$ represents the coupling of $(X_0, X_1)$, $p(t)$ denotes the time distribution defined on $[0,1]$. During the inference process, the ODE usually needs to be discretized and solved via the Euler method, expressed as:
\begin{equation}
{X_{t + \Delta t}} = {X_t} + \Delta t \cdot {v_\theta }({X_t},t),t \in \left\{ {0,\Delta t, \ldots ,(N - 1) \Delta t} \right\},
\label{eq:3}
\end{equation}
where $\Delta t = \frac{1}{N}$, $N$ represents the total number of steps. Generally, a larger $N$ results in higher-quality generated images, but it requires more computational resources for sampling.
The reference trajectory of RF is a straight flow. If RF can converge to the straight flow through training, it can significantly minimize numerical errors in the ODE solver, enabling few-step or single-step generation: as the state moves along the trajectory with uniform linear motion, the velocity at a single time step equals the average velocity over the entire trajectory. 
As long as the cumulative sum of all time increments throughout the motion equals unity, the final endpoint of the trajectory will remain consistent.

\subsection{Analysis of Reflow}
\label{3.2}
The inference trajectory of RF exhibits a curved pattern when trained using the vanilla method. While RF attributes this phenomenon to the crossing of reference trajectories \cite{liu2022rectified} and proposes Reflow as a solution (with implementation details provided in Appendix C), our experiments (Fig.\ref{reuse}) and analysis suggest that the effectiveness of Reflow cannot be solely explained by its ability to mitigate trajectory-crossing issues. This is particularly evident given that trajectory crossings are extremely rare in high-dimensional spaces, as we will elaborate in Section \ref{4.1}. 


\begin{figure}[t!]
\centering
\centerline{\includegraphics[width=\columnwidth]{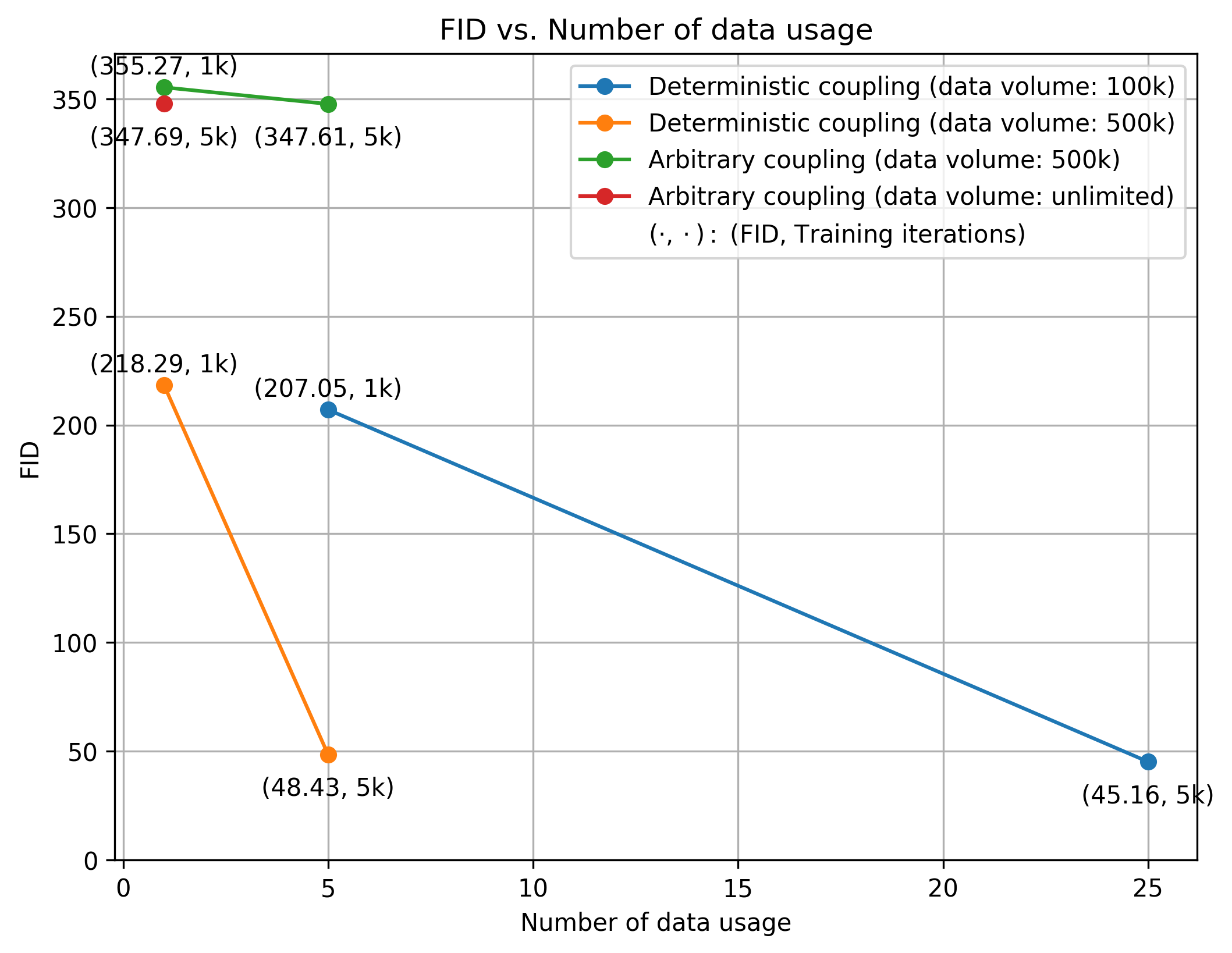}}
\caption{\textbf{Impact of coupling type and data reuse on RF's performance.} Training RF from scratch with different coupling types, we control the data volume to construct different data reuse scenarios and observe the impact of coupling type and data reuse on model performance under the same training iterations. Each iteration samples 500 data pairs, ensuring that the total data volume seen by the model remains the same across different reuse scenarios under the same training iterations.}
\label{reuse}
\end{figure}

Through experimental and analytical investigations, the success of Reflow can also be attributed to the following two factors: (1) the images in the training data for Reflow are generated by the pre-trained model, i.e., \textit{deterministic couplings}. This implies the existence of a deterministic trajectory between a coupling that can be learned and inferred by the model, as shown in the first row of Fig.\ref{dongji}. Training/fine-tuning with Reflow is analogous to learning this trajectory during training and progressively aligning it with a straight-line trajectory. In contrast, randomly matched data pairs (\textit{arbitrary couplings}) lack explicit trajectory relationships, requiring the model to infer unknown trajectories from noises and constrain them to straight lines, making the training more complex and unstable. As demonstrated in Fig.\ref{reuse}, under identical data volume and training iterations, RF trained with \textit{deterministic couplings} achieves superior image quality (green line vs. orange line). (2) Reflow reuses data by sampling intermediate states from different time steps of the same trajectory for training, as depicted in the first row of Fig.\ref{dongji}. This training strategy resembles the multi-time-scale optimization methods employed in the distillation, both of which aim to enhance model performance by leveraging multi-level information. Fig.\ref{reuse} demonstrates that more data reuse iterations lead to better performance under \textit{deterministic couplings} (orange line vs. blue line). More details and analyses can be found in Appendix B.

Additionally, the drawbacks of Reflow are evident. As shown in Fig.\ref{dongji}, there is a distribution gap between the generated images and the real images. Each iteration of training utilizes images generated by the previous model as training data, leading to the accumulation of errors over time. 
Consequently, the quality of the images generated by the subsequent model tends to degrade compared to the previous one. To balance high image quality with computational efficiency, it is typically recommended to limit the iterative training to 2 or 3 cycles.
Furthermore, while data reuse is effective, its effectiveness is influenced by the volume of data. If the volume of reused data is small, it may not adequately represent the true probabilistic distribution of the data, thereby affecting the model's learning of the PF \cite{liu2022flow}. On the other hand, excessive reused data will cause significant storage pressure, which increases the consumption of computational resources and reduces training efficiency.


\section{Method}

\begin{figure*}[t!]
\centering
\centerline{\includegraphics[width=\textwidth]{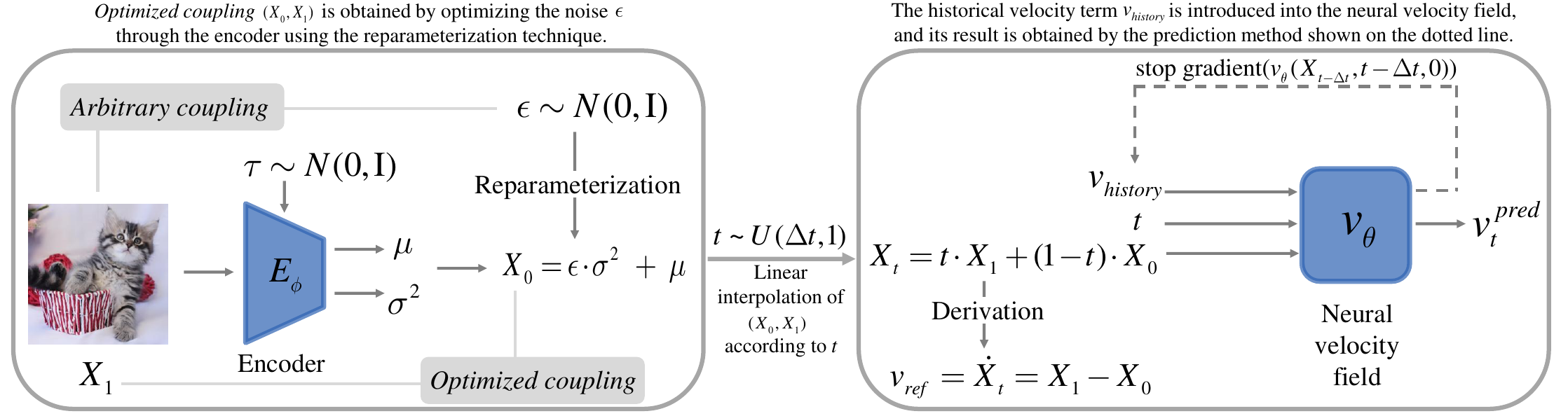}}
\caption{\textbf{Overview of VRFNO.} The encoder and the neural velocity field form a joint training framework: randomly matched noise $\epsilon$ and image $X_1$ (called \textit{arbitrary coupling}) are optimized by the encoder to obtain \textit{optimized coupling} $(X_0,X_1)$, which are then used to train the neural velocity field. The introduction of the HVT $v_{history}$ enhances the distinction of the trajectories. Both components work together during training to straighten the inference trajectories, thereby improving the model's predictive accuracy.}
\label{fangfa}
\end{figure*}
We aim to develop an advanced training framework for RF as a superior alternative to Reflow, further improving the quality of generated images in single-step and few-step generation tasks. To achieve this, we propose a novel method called VRFNO and the overview of VRFNO is shown in Fig.\ref{fangfa}.
Specifically, VRFNO introduces an HVT to effectively improves the model's capability to differentiate inference trajectories, thereby substantially mitigating prediction disturbances arising from intermediate state similarities.
Furthermore, to circumvent the use of pre-trained model-generated images as training data during the training process, as employed in Reflow, and to ensure that randomly matched noise-image data pairs $(X_0, X_1)$ exhibit desirable characteristics akin to deterministic coupling. We define a new concept “\textit{optimized coupling}” and propose to achieve \textit{optimized coupling} by optimizing noise.
\begin{definition}
$(X_0, X_1)$ is called an optimized coupling if it satisfies the following condition:
\[
\left\| {{v_\theta }(t{X_1} + (1 - t){X_0}) - ({X_1} - {X_0})} \right\| \le  \varepsilon,
\]
where $\varepsilon > 0$ is a small positive constant.
\end{definition}

\subsection{Viscous Rectified Flow}
\label{4.1}
RF attributes the curvature of inference trajectories to the frequent crossings of reference trajectories. Specifically, when noise-image pairs are randomly matched, multiple reference trajectories may cross at the same point $X_t$ at time step $t$. During training, this causes different reference trajectories to provide identical inputs to the model at time step $t$, confusing the model and degrading its prediction accuracy.
However, our calculations (Theorem \ref{Theorem_1}) show that the probability of trajectory crossings occurring during actual training is $P \sim O(e^{-c(n \times n)})$, which is extremely rare in high-dimensional space. 
More commonly, since the initial states are independently sampled from a standard Gaussian distribution, their statistical properties are very similar. In the early stages of the interpolation trajectory, the intermediate states contain limited image information, causing the differences between the intermediate states to remain small. Since RF learns the mapping relationship between distributions, this high similarity in statistical properties may hinder the model's ability to distinguish between states, leading to blurred predictions. Ideally, RF has a constant velocity field, which means that the intermediate state ${X_t}$ moves in a straight line from noise $X_0$ to image $X_1$ at uniform velocity $v_{ref}$, forming a straight inference trajectory.
\begin{theorem}
\label{Theorem_1}
In $(n \times n)$-dimensional space, for straight-line interpolation trajectories $X^{(\cdot)} = \left \{ X_t^{(\cdot)} : t \in  [0, 1] \right \} $, the probability of $X^{(i)}$ and $X^{(j)}$ crossing at the point $X_t$ at time step $t$ is $P \sim O(e^{-c(n \times n)}), c > 0$.
\end{theorem}
As shown in Theorem \ref{Theorem_2}, the velocity difference $\Delta ({v_{ref}}^{(i)},{v_{ref}}^{(j)})$ between each $X_t$ is more significant than their state difference $\Delta ({X_t}^{(i)},{X_t}^{(j)})$, and this more significant difference enables the model to better identify the currently inferred trajectory, thus improving prediction accuracy. Therefore, we introduce an HVT $v_{history}$ as an approximation of the constant velocity, which is fed into the model at each time step, resulting in the Viscous Rectified Flow (VRF):
\begin{equation}
d{X_t} = v({X_t},t,{v_{history}})dt,~~t \in [\Delta t,1],
\label{eq:4}
\end{equation}
where $\Delta t$ is the time interval between the current state and the adjacent history state. During the inference process, the HVT input to the model at the first time step is 0. In subsequent sampling iterations, each iteration inputs the velocity predicted by the model at the previous time step into the model as an HVT, providing the model with auxiliary information on the flow direction and enhancing the model's ability to make correct predictions.

\begin{theorem}
\label{Theorem_2}
For each intermediate state $X_t$ along a PF trajectory, the velocity difference between $X_t$ are greater than their state difference:
\[\Delta ({v_{ref}}^{(i)},{v_{ref}}^{(j)}) \ge \Delta ({X_t}^{(i)},{X_t}^{(j)}),\]
where $\Delta(\cdot,\cdot)=\mathbb{E} [ \left \| \cdot-\cdot \right \|_F^2 ] $ and $\mathbb{E}[\cdot]$ is the expectation.
\end{theorem}

\begin{algorithm}[tb]
   \caption{Training of VRFNO.}
   \label{alg:1}
\begin{algorithmic}
   \STATE {\bfseries Input:} image $X_1$, noise $\epsilon$, learning rate $\eta $, time interval $\Delta t$, hyperparameter $\alpha$, encoder $E_\phi$, velocity field $v_\theta$.
   \REPEAT
   \STATE $\mu, \sigma^2 = E_\phi(X_1)$
   \STATE $X_0= \epsilon \cdot {\sigma ^2}+\mu$
   \STATE $v_{ref}=X_1-X_0$
   \STATE Sample $t \sim U(\Delta t,1)$
   \STATE ${X_t} = t{X_1} + (1 - t){X_0}$
   \STATE ${X_{t - \Delta t}} = (t - \Delta t){X_1} + (1 - t + \Delta t){X_0}$
   \STATE ${v_{history}} = stopgrad({v_\theta }({X_{t - \Delta t}},t - \Delta t,0))$
   \STATE $v_t = {v_\theta }(X_t,t,v_{history})$
   \STATE $L(\theta,\phi)\gets$
   \STATE $ ~~~~\mathbb{E}{\left [ d( v_{ref} , {v_t } ) \right ] } + \frac{\alpha}{2}{\left( {\sigma^2 + \mu^2 - 1 - \log (\sigma^2)} \right)}  $
   \STATE $\theta \gets \theta - \eta\nabla_\theta L(\theta,\phi)$
   \STATE $\phi \gets \phi - \eta\nabla_\phi L(\theta,\phi)$
   \UNTIL{convergence}
\end{algorithmic}
\end{algorithm}

\subsection{Noise Optimization for Optimized Coupling}
We propose using noise reparameterization technique to optimize the noise representation. Specifically, we design a joint training framework for an encoder and a neural velocity field (as shown in Fig.\ref{fangfa}), where the encoder optimizes the noises to achieve \textit{optimized couplings} and uses these optimized data pairs to train the neural velocity field.
The complete training process is shown in Algorithm \ref{alg:1}.
We adopt the working mode of the encoder in the VAE architecture \cite{higgins2017beta}: the encoder $E_\phi$ takes the image $X_1$ as input and outputs the corresponding mean matrix $\mu$ and variance matrix $\sigma^2$, then generates the optimized noise $X_0$ by combining $\mu$ and $\sigma^2$ with randomly matched noise $\epsilon$ using the reparameterization technique:
\begin{equation}
{X_0} = \epsilon \cdot {\sigma ^2}+\mu.
\label{eq:5}
\end{equation}
Given the limited number of images in the dataset, to prevent the model from developing a memory effect and ensure diversity in generated images, we introduce random perturbation $\tau \sim N(0,\mathrm{I})$ at the intermediate layers of the encoder, as shown in Fig.\ref{fangfa}. This ensures that the reparameterized mean and variance matrices still exhibit differences even when multiple Gaussian noises are matched to the same image. Furthermore, after these Gaussian noises matched the same image are reparameterized, they tend to concentrate in a smaller subspace. During training, each image is matched with noise from its corresponding subspace and participates in the training. This approach is similar to the data reuse strategy in Reflow, and it avoids the issue of limiting the model's generalization ability due to the reuse of limited data, thereby helping to improve the performance of the model.
After obtaining the optimized noise $X_0$, we perform straight-line interpolation between it and the corresponding image $X_1$ to obtain $X_t$ and $X_{t-\Delta t}$:
\begin{equation}
\begin{array}{*{20}{c}}
{{X_t} = t{X_1} + (1 - t){X_0}}\\
{{X_{t - \Delta t}} = (t - \Delta t){X_1} + (1 - t + \Delta t){X_0}}
\end{array},{\rm{  }}~~t \in [\Delta t ,1].
\label{eq:6}
\end{equation}
We use the $stopgrad$ operator to calculate the HVT ${v_{history}} = stopgrad({v_\theta }({X_{t - \Delta t}},t - \Delta t,0))$. 
The neural velocity field $v_\theta$ takes the intermediate state $X_t$, time step $t$, and HVT as inputs, and learns a constant velocity field by fitting the reference velocity $v_{ref}$:
\begin{equation}
VCL(\theta ,\phi ) = {\mathbb{E} _{t \in p(t)}}\left[ {{d( {v_{ref} , {v_\theta }({X_t},t,{v_{history}})}) }} \right],
\label{eq:7}
\end{equation}
where $d(\cdot,\cdot)$ indicates distance metric. This process simultaneously optimizes the encoder $E_\phi$ to ensure that the reparameterized noises and the corresponding images satisfy the property of \textit{optimized couplings}.
Meanwhile, to prevent the encoder from overfitting during training, which could cause the reparameterized noises to rely too heavily on image information from datasets and thus constrain the diversity of generated images, we introduce KL divergence regularization to constrain the mean and variance of the encoder's output near the standard Gaussian distribution:
\begin{equation}
KLL(\phi ) = \frac{1}{2}{\left( {\sigma^2 + \mu^2 - 1 - \log (\sigma^2)} \right)}.
\label{eq:8}
\end{equation}
Therefore, the total loss of our joint training framework is:
\begin{equation}
L(\theta ,\phi ) = VCL(\theta ,\phi ) + \alpha KLL(\phi ),
\label{eq:8}
\end{equation}
where $\alpha$ is a hyperparameter that controls the strength of regularization.
By using the \textit{optimized couplings} optimized through the encoder to train the neural velocity field, it can gradually learn how to infer nearly straight trajectories, thereby enabling few-step and single-step sampling. Furthermore, our encoder-neural velocity field joint generation framework still satisfies the marginal preserving property (Theorem \ref{Theorem_3}), with a detailed definition and proof provided in the Appendix A.

\begin{theorem}
\label{Theorem_3}
Assume $X$ is rectifiable and $Z$ is its viscous rectified flow. The marginal law of $Z_t$ equals that of $X_t$ at every time $t$, i.e., Law($Z_t$) = Law($X_t$), $\forall t \in [0,1]$.
\end{theorem}

\begin{algorithm}[t]
   \caption{Sampling of VRFNO.}
   \label{alg:2}
\begin{algorithmic}
   \STATE {\bfseries Input:} image in dataset $X_1$, sampling steps $N$, encoder $E_\phi$, velocity field $v_\theta$.
   \STATE {\bfseries Output:} generated image $\tilde{X} _1$.
   \STATE Sample $\epsilon \sim N(0,\mathrm{I})$
   \STATE $\mu, \sigma^2 = E_\phi(X_1)$
   \STATE $X_0= \epsilon \cdot {\sigma ^2}+\mu$
   \STATE $\Delta t = 1/N$
   \STATE $v_{history}=0$
   \FOR{$i=0 $ {\bfseries to} $N-1$}
      \STATE $v_{pred}=v_\theta(X_{i\Delta t},i\Delta t, v_{history})$
      \STATE $v_{history}=v_{pred}$
      \STATE $X_{(i+1)\Delta t}=X_{i\Delta t} + \Delta t \cdot v_{pred}$
   \ENDFOR
   \STATE $\tilde{X} _1 = X_{N\Delta t}$
\end{algorithmic}
\end{algorithm}

\noindent\textbf{Noise optimization.}  
In previous studies, noise optimization typically treats noise as a learnable parameter and performs small-scale iterative updates over multiple iterations using gradient information provided by the model, until it converges to an approximate optimal solution \cite{eyring2025reno, zhou2024golden}. While it can generate high-quality images, its limitations are also evident: it requires personalized optimization for each noise and relies on pre-trained models.
In contrast, our method utilizes the reparameterization technique to optimize the noise through linear transformations directly. This approach allows for significant adjustments to the noise in a single optimization step, enabling rapid and efficient approximation to the optimal solution, significantly reducing the number of iterations and computational overhead, while also eliminating the dependence on pre-trained models.

\noindent\textbf{Two-stage training.} In the training process, we adopt a two-stage strategy based on empirical observations. In the first stage, we set $d(\cdot,\cdot)$ to the MSE loss, and once convergence is achieved, we proceed to the second stage. In this stage, we incorporate the LPIPS loss \cite{zhang2018unreasonable} for joint training on top of the first-stage setup, continuing until convergence is reached again. Notably, our method does not rely on distillation or adversarial training mechanisms.

\noindent\textbf{Sampling.} After training the encoder and the neural velocity field, we can generate images using VRF ODE introduced in Eq.(\ref{eq:4}), the discrete sampling process is shown in Algorithm \ref{alg:2}. For single-step generation, the HVT is set to 0 during sampling, while in few-step generation, the HVT is updated at each step. Additionally, we integrate the dataset into the sampler, meaning that the sampled noise is also reparameterized during the generation process. Since we introduced perturbations in the encoder, this will not affect the diversity of the generated images but can further enhance the quality of the generated images.

\section{Experiment}
We evaluate the performance of VRFNO in various scenarios, including synthetic data and real data. In Section \ref{5.1}, we compare the performance of RF and VRFNO by studying the inference trajectories between two 2D Gaussian distributions, which clearly demonstrates the effectiveness of our method.  In Section \ref{5.2}, we extend the experiments to real image datasets, CIFAR-10 \cite{krizhevsky2009learning} and AFHQ \cite{choi2020stargan}, at different resolutions to validate the performance of VRFNO in one-step and few-step generation. Additionally, we conducted further analysis on VRFNO, including ablation studies, as well as quantitative evaluations of its trajectory straightness and inference efficiency.

\noindent\textbf{Baselines and evaluation.} We evaluated VRFNO and compared it with state-of-the-art diffusion models \cite{song2020score, luhman2021knowledge, salimans2022progressive, song2023consistency, karras2022elucidating} and rectified flow models (RF \cite{liu2022flow}, Constant Acceleration Flow (CAF) \cite{park2024constant}, TraFlow \cite{wu2025traflowtrajectorydistillationpretrained}, Shortcut Model (SM) \cite{frans2024one}) in single-step and few-step generation. Among them, RF serves as the baseline for our method, while CAF is the most similar approach to ours. Therefore, these two methods are the primary focus of our comparison. We use Fréchet Inception Distance (FID) and Kernel Inception Distance (KID) to assess the quality of generated images, and use Inception Score (IS) to measure diversity. Lower FID and KID indicate higher generation quality, while a higher IS suggests greater diversity in the generated images.

\begin{figure}[t]
\centering
\centerline{\includegraphics[width=\columnwidth]{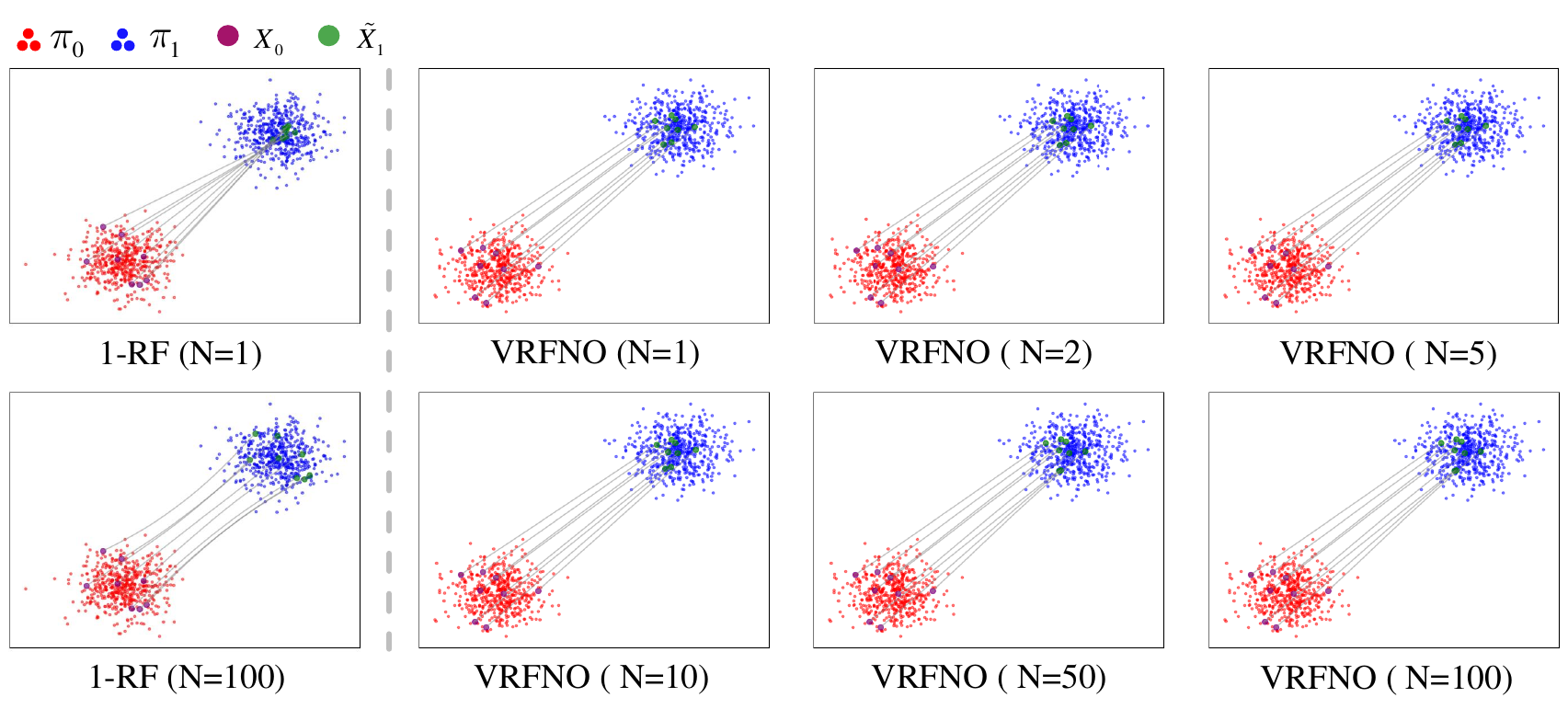}}
\caption{\textbf{Visualization of inference trajectories on synthetic data.} The first column on the left shows the inference trajectories of 1-RF with $N$ steps, and the three columns on the right show the inference trajectories of VRFNO with $N$ steps.}
\label{toy}
\end{figure}
\begin{table}[]
\centering
\caption{Quantitative results on CIFAR-10.}
\label{cifar10}
\begin{tabular}{ccccc}
\hline
Methods &
  \begin{tabular}[c]{@{}c@{}}NFE\end{tabular} &
  \begin{tabular}[c]{@{}c@{}}IS\\ ($\uparrow$)\end{tabular} &
  \begin{tabular}[c]{@{}c@{}}FID\\ ($\downarrow$)\end{tabular} &
  \begin{tabular}[c]{@{}c@{}}KID($\times 10^{-3}$)\\ ($\downarrow$)\end{tabular} \\ \hline
\multicolumn{5}{l}{Diffusion/Consistency Models} \\ \hline
\begin{tabular}[c]{@{}c@{}}VP ODE\\ (+distill)\end{tabular} &
  1 &
  \begin{tabular}[c]{@{}c@{}}1.20\\ (8.73)\end{tabular} &
  \begin{tabular}[c]{@{}c@{}}451\\ 16.23)\end{tabular} &
  - \\
\begin{tabular}[c]{@{}c@{}}sub-VP ODE\\ (+distill)\end{tabular} &
  1 &
  \begin{tabular}[c]{@{}c@{}}1.21\\ (8.80)\end{tabular} &
  \begin{tabular}[c]{@{}c@{}}451\\ (14.32)\end{tabular} &
  - \\
DDIM distillation  & 1   & 8.36  & 9.36  & -     \\
Progressive        & 1   & -     & 9.12  & -     \\
CT                 & 1   & 8.49  & 8.71  & -     \\
EDM                & 5   & -     & 37.75 & -     \\ \hline
\multicolumn{5}{l}{Rectified Flow Models}        \\ \hline
1-RF               & 1   & 1.13  & 379   & 428   \\
2-RF               & 1   & 8.15  & 11.97 & 8.66  \\
CAF                & 1   & 8.32  & 4.81  & -     \\
TraFlow            & 1   & -     & 4.50  & -     \\
SM                 & 1   & -     & 4.93  & -     \\
VRFNO (Ours)         & 1   & \textbf{9.59} & \textbf{4.50}  & \textbf{2.73}  \\ \hline
1-RF               & 5   & 7.12  & 34.81  & 32.26  \\
2-RF               & 5   & 9.01  & 4.36  & 2.25  \\
CAF                & 5   & 8.67  & 4.03  & -     \\
VRFNO (Ours)         & 5   & \textbf{9.42}  & \textbf{4.03} & \textbf{2.13} \\ \hline
1-RF               & 10  & 8.44  & 12.70  & 11.50  \\
2-RF               & 10  & 9.13  & 3.83  & 1.63  \\
CAF                & 10  & 9.12  & 3.77  & -     \\
VRFNO (Ours)         & 10  & \textbf{9.51}  & \textbf{3.36}  & \textbf{1.31}  \\ \hline
\end{tabular}
\end{table}
\subsection{Synthetic Experiments}
\label{5.1}
We demonstrate the trajectory straightening effect of VRFNO through 2D synthetic data experiments.
For the neural velocity field, we use a simple multi-layer perceptron (MLP) architecture with three hidden layers, each containing 64 units. For the encoder, we adopt the encoder architecture from the VAE framework, consisting of two hidden layers, each with 16 units. Additionally, a fully connected layer is used to produce the mean and variance separately.

The visualization results are shown in Fig.\ref{toy}, where we compared VRFNO with the 1st generation RF (1-RF). Additionally, in Appendix B, we provide the inference trajectories of 1-RF and the 2nd generation RF (2-RF) for different numbers of steps. In the multi-step generation, the inference trajectory of VRFNO is closer to a straight line compared to 1-RF, which provides a guarantee for accurate single-step and few-step generation. The single-step and few-step inference trajectories of VRFNO are visually straight, indicating that the neural velocity field of VRFNO better converges to a constant velocity field, making its predictions of the trajectory direction more stable. In addition, the error between the endpoints of the single-step and the few-step inference trajectories and the endpoints of the multi-step inference trajectories is small, indicating that VRFNO is able to accurately map samples from the initial distribution to the target distribution in both single-step and few-step generation.
In contrast, when 1-RF is generated in a single step, the endpoint of the trajectory is usually concentrated at the mean point of the target distribution, while the samples near this point are usually meaningless in practical applications, as shown in Fig.$\mathrm{II}$ (a) in Appendix B.

\subsection{Real-data Experiments}
\label{5.2}

To further validate the effectiveness of our method, we evaluate VRFNO on real image datasets, namely CIFAR-10 with a resolution of $32 \times 32$ and AFHQ with different resolutions (including $64 \times 64$, $128 \times 128$, $256 \times 256$). For the neural velocity field, we adopt the UNet architecture from DDPM++ \cite{song2020score} to train on CIFAR-10 and the NCSNv2 \cite{song2020improved} to train on AFHQ; for the encoder, we refer to the encoder architecture from \cite{higgins2017beta}, and its details can be found in Appendix E. Specifically, the encoder requires very few parameters, taking the train of CIFAR-10 as an example, the number of parameters in the encoder is less than $1/20$ of those in the neural velocity field.
\begin{figure}[t!]
\centering
\centerline{\includegraphics[width=\columnwidth]{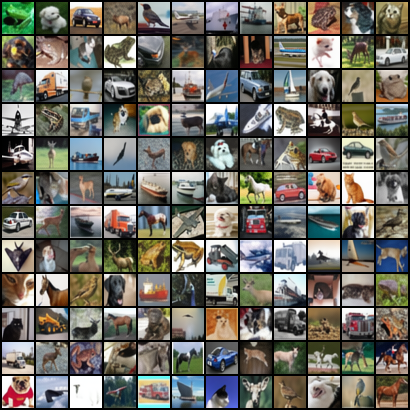}}
\caption{\textbf{Qualitative results on CIFAR-10.} Visualization of one-step generation of VRFNO}
\label{cifar_1step}
\end{figure}

\begin{figure}[t!]
\centering
\centerline{\includegraphics[width=\columnwidth]{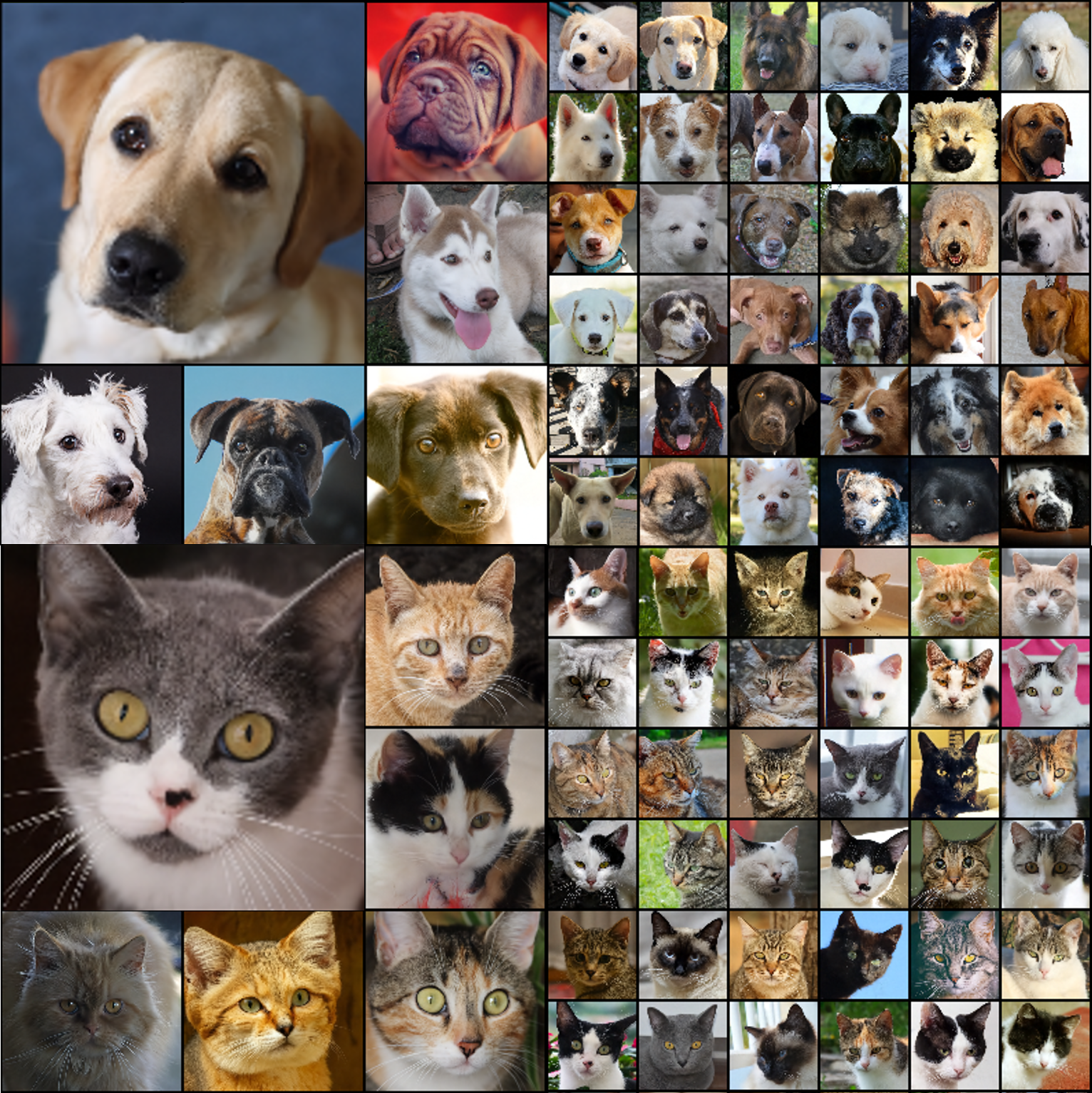}}
\caption{\textbf{Qualitative results on AFHQ.} Visualization of one-step generation of VRFNO at different resolutions}
\label{afhq_1step}
\end{figure}
\begin{table}
\centering
\caption{Quantitative results on AFHQ with different resolutions.}
\label{afhq}
\begin{tabular}{ccccc}
\hline
Methods & Datasets                                                             & NFE                & Resolution     & FID($\downarrow$) \\ \hline
\multirow{6}{*}{\begin{tabular}[c]{@{}c@{}}2-RF\\ (+distill)\end{tabular}} &
  \multirow{3}{*}{\begin{tabular}[c]{@{}c@{}}AFHQ\\ -CAT\end{tabular}} &
  \multirow{3}{*}{1} &
  $64\times64$ &
  181.93(31.57) \\
        &                                                                      &                    & $128\times128$ & 172.66(29.81)     \\
        &                                                                      &                    & $256\times256$ & 171.84(29.33)     \\ \cline{2-5} 
        & \multirow{3}{*}{\begin{tabular}[c]{@{}c@{}}AFHQ\\ -DOG\end{tabular}} & \multirow{3}{*}{1} & $64\times64$   & 200.77(33.64)     \\
        &                                                                      &                    & $128\times128$ & 192.30(32.70)     \\
        &                                                                      &                    & $256\times256$ & 189.82(30.27)     \\ \hline
\multirow{6}{*}{\begin{tabular}[c]{@{}c@{}}VRFNO\\ (Ours)\end{tabular}} &
  \multirow{3}{*}{\begin{tabular}[c]{@{}c@{}}AFHQ\\ -CAT\end{tabular}} &
  \multirow{3}{*}{1} &
  $64\times64$ &
  28.69 \\
        &                                                                      &                    & $128\times128$ & 27.56             \\
        &                                                                      &                    & $256\times256$ & 27.04             \\ \cline{2-5} 
        & \multirow{3}{*}{\begin{tabular}[c]{@{}c@{}}AFHQ\\ -DOG\end{tabular}} & \multirow{3}{*}{1} & $64\times64$   & 44.64             \\
        &                                                                      &                    & $128\times128$ & 27.21             \\
        &                                                                      &                    & $256\times256$ & 27.37             \\ \hline
\end{tabular}
\end{table}

Tab.\ref{cifar10} presents the quantitative evaluation of VRFNO on CIFAR-10. Compared to other diffusion models, our approach achieves superior performance in single-step generation, with FID = 4.53, KID = 2.73, and IS = 10.59. Compared to previous rectified flow models, VRFNO attains state-of-the-art performance in both single-step and few-step generation (our training does not involve distillation).
Fig.\ref{cifar_1step} illustrates the qualitative evaluation of VRFNO in single-step generation on CIFAR-10. Further comparative qualitative evaluations between VRFNO and RF in N-step generation can be found in Appendix B.
Tab.\ref{afhq} provides the quantitative evaluation of VRFNO's generated images on the AFHQ cat and dog datasets, focusing on the image quality of single-step generation at different resolutions. Our method is designed as a better alternative to Reflow, so the comparison primarily highlights the performance difference with 2-RF. From the tab.\ref{afhq}, it is evident that VRFNO outperforms 2-RF in image generation quality across different resolutions.The qualitative evaluation of AFHQ at different resolutions can be found in Fig.\ref{afhq_1step}.

\subsection{Further Analysis}
\noindent\textbf{Ablation Study.} 
We conduct an ablation study to evaluate the effectiveness of each component in our framework for single-step and few-step generation. Specifically, we investigate the improvements brought to RF by (1) the introduction of the HVT and (2) the noise optimization operation. The configurations and results are shown in Tab.\ref{ablation}.
The two components have both contributed to the improvement of model performance. Regarding the introduction of the HVT, even with one-step sampling, the model performance improved, indicating that it facilitates trajectory rectification during the training process. As for the noise optimization operation, it results in a significant performance boost, demonstrating the effectiveness of training the neural velocity field with our defined \textit{optimized coupling}.

\begin{table}[]
\centering
\caption{Ablation study of HVT and noise optimization.}
\label{ablation}
\begin{tabular}{ccccc}
\hline
Config & \begin{tabular}[c]{@{}c@{}}NFE\end{tabular} & HVT & \begin{tabular}[c]{@{}c@{}}Noise \\ optimization\end{tabular} & FID($\downarrow$) \\ \hline
A & 1  & \ding{55} & \ding{55} & 379 \\
B & 1  & \ding{51} & \ding{55} & 332 \\
C & 1  & \ding{55} & \ding{51} & 4.72 \\
D & 1  & \ding{51} & \ding{51} & 4.53 \\\cline{2-5}
E & 5  & \ding{55} & \ding{55} & 34.81 \\
F & 5  & \ding{51} & \ding{55} & 32.50 \\
G & 5  & \ding{55} & \ding{51} & 4.28 \\
H & 5  & \ding{51} & \ding{51} & 4.03 \\\cline{2-5}
I & 10 & \ding{55} & \ding{55} & 12.70 \\
J & 10 & \ding{51} & \ding{55} & 9.34 \\
K & 10 & \ding{55} & \ding{51} & 4.75 \\
L & 10 & \ding{51} & \ding{51} & 3.40 \\ \hline
\end{tabular}
\end{table}

\begin{table}[]
\centering
\caption{Comparison of flow straightness.}
\label{straightness}
\begin{tabular}{ccccc}
\hline
Dataset  & 2-RF  & 3-RF  & CAF   & VRFNO(Ours) \\ \hline
2D       & 0.067 & 0.053 & 0.058 & 0.054 \\
CIFAR-10 & 0.058 & 0.056 & 0.035 & 0.026 \\ \hline
\end{tabular}
\end{table}

\begin{table}[]
\centering
\caption{Comparison of the inference time (sec) in $N$-step.}
\label{speed}
\begin{tabular}{cccc}
\hline
NFE & RF    & CAF   & VRFNO(Ours) \\ \hline
1          & 0.172 & 0.181 & 0.305       \\
10         & 1.404 & 1.415 & 1.646       \\ \hline
\end{tabular}
\end{table}

\noindent\textbf{Flow Straightness.}
To evaluate the straightness of the inference trajectories, we introduce the Normalized Flow Straightness Score (NFSS) \cite{liu2022flow,liu2023instaflow}. A smaller NFSS indicates that the inference trajectory is closer to a straight line. 
We compare VRFNO with RF and CAF on synthetic datasets and CIFAR-10, with the experimental results shown in Tab.\ref{straightness}. Our method generates inference trajectories that are straighter compared to the other methods.

\noindent\textbf{Time Efficiency.}
We also compare the inference time of our method with RF and CAF. Tab.\ref{speed} presents the time required to generate 512 images in both one-step and ten-step settings. Our method incurs a slightly higher time cost than RF due to the additional encoder. In contrast, CAF consists of two neural velocity fields, it necessitates an additional computation (its total NFE equals $N+1$), resulting in significantly higher time consumption.

\section{Related Work}

The inference process of DMs can be viewed as an iterative solution of ODE \cite{song2020denoising} or stochastic differential equations \cite{bao2022analytic, ho2020denoising}, and ODE-based methods \cite{zhang2022fast, dockhorn2022genie} are more effective in few-step sampling due to their determinism.  
A widely used first-order ODE solver is the Euler solver, but its large local truncation error necessitates more iteration steps for high-quality image generation.
To reduce iteration steps, higher order ODE solvers such as DEIS \cite{zhang2022fast} and DPM solvers \cite{lu2022dpm} exploit the semilinear structure of diffusion ODEs, deriving exact solutions for high-quality image generation with fewer steps.
DPM-solver++ \cite{lu2022dpmp} enhances DPM-solver by improving stability in higher-order solvers.
UniPC \cite{zhao2024unipc} builds a unified predictor-corrector framework for DPM solver, enhancing the quality of sampling in a few steps.
DC-solver \cite{zhao2024dc} uses dynamic compensation to correct misalignment in UniPC, thus improving image quality.

Another mainstream method for accelerated sampling is distillation.
Progressive distillation \cite{salimans2022progressive} iteratively halves the steps of a DDIM until it enables image generation in just 4 steps.
Diff-Instruct \cite{luo2024diff} accelerates sampling by minimizing Integra Kullback-Leibler divergence to merge knowledge from multiple time steps of a pre-trained model.
Consistency Model (CM) \cite{song2023consistency} enables direct mapping from noise to data by minimizing the differences between the final states when adjacent ODE trajectory points are used as inputs.
Latent CM \cite{luo2023latent} treats the guided reverse diffusion process as solving an augmented PF ODE, directly predicting the ODE solution in the latent space, enabling fast and high-fidelity sampling.
Consistency Trajectory Model \cite{kim2023consistency} further optimizes CM by allowing the model to learn mappings between arbitrary initial and final times along the ODE trajectory during the diffusion process.  

\section{Conclusion}
This paper proposes a novel method VRFNO, which is a joint training framework combining an encoder (for noise optimization) and a neural velocity field (for velocity prediction). VRFNO utilizes the encoder to optimize noises to transform randomly matched noises and images into \textit{optimized couplings}, which are then used to train the neural velocity field for precise prediction of straight inference trajectories. To further straighten the inference trajectories during training, we introduce the HVT into the neural velocity field to enhance its prediction accuracy. Extensive experiments conducted on both synthetic and real world image datasets demonstrate the effectiveness and scalability of our approach. The empirical results show that VRFNO achieves state-of-the-art performance across multiple datasets and varying resolutions.
{
    \small
    \bibliographystyle{ieeenat_fullname}
    \bibliography{main}
}

\end{document}